\documentclass[lettersize,journal]{IEEEtran}
\usepackage{amsmath,amsfonts}
\usepackage{algorithmic}
\usepackage{algorithm}
\usepackage{array}
\usepackage[caption=false,font=normalsize,labelfont=sf,textfont=sf]{subfig}
\usepackage{textcomp}
\usepackage{stfloats}
\usepackage{url}
\usepackage{verbatim}
\usepackage{graphicx}
\usepackage{cite}
\usepackage{booktabs}
\usepackage{multirow}
\usepackage{enumitem}

\usepackage{threeparttable}
\usepackage{setspace}
\usepackage{colortbl}
\definecolor{mygray2}{gray}{.6}
\definecolor{mygray3}{gray}{.3}

\usepackage[capitalize]{cleveref}
\crefname{section}{Sec.}{Secs.}
\Crefname{section}{Section}{Sections}
\crefname{table}{Tab.}{Tabs.}
\Crefname{table}{Table}{Tables}

\def\eg{\emph{e.g}\@addpunct{.}}
\def\eg{\emph{e.g}\@addpunct{.}}
\def\Eg{\emph{E.g}\@addpunct{.}}
\def\ie{\emph{i.e}\@addpunct{.}} 
\def\Ie{\emph{I.e}\@addpunct{.}}
\def\cf{\emph{cf}\@addpunct{.}} 
\def\Cf{\emph{Cf}\@addpunct{.}}
\def\etc{\emph{etc}\@addpunct{.}} 
\def\vs{\emph{vs}\@addpunct{.}}
\def\wrt{w.r.t\@addpunct{.}} 
\def\dof{d.o.f\@addpunct{.}}
\def\iid{i.i.d\@addpunct{.}} 
\def\wolog{w.l.o.g\@addpunct{.}}
\def\etal{\emph{et al}\@addpunct{.}}

\begin{document}

\title{Context-Enhanced Video Moment Retrieval with Large Language Models}


\author{
Weijia~Liu, Bo~Miao, Jiuxin~Cao, Xuelin~Zhu, Bo~Liu, Mehwish~Nasim, Ajmal~Mian
\IEEEcompsocitemizethanks{
\IEEEcompsocthanksitem W.~Liu, J.~Cao and X.~Zhu are with School of Cyber Science and Engineering, Southeast University, Nanjing 211189, China.\protect\\
(e-mail: weijia-liu@seu.edu.cn, jx.cao@seu.edu.cn, zhuxuelin@seu.edu.cn)\\
(Jiuxin.Cao is the corresponding author.)
\IEEEcompsocthanksitem B.~Miao, M.~Nasim and A.~Mian are with School of Computer Science and Software Engineering, University of Western Australia, Perth 6009, Australia.\\
(e-mail: bo.miao@research.uwa.edu.au, mehwish.nasim@uwa.edu.au, ajmal.mian@uwa.edu.au)
\IEEEcompsocthanksitem B. Liu is with School of Computer Science and Engineering, Southeast University, Nanjing 211189, China.\\
(e-mail: bliu@seu.edu.cn)
}
}



\maketitle

\begin{abstract}
Current methods for Video Moment Retrieval (VMR) struggle to align complex situations involving specific environmental details, character descriptions, and action narratives.
To tackle this issue, we propose a Large Language Model-guided Moment Retrieval (LMR) approach that employs the extensive knowledge of Large Language Models (LLMs) to improve video context representation as well as cross-modal alignment, facilitating accurate localization of target moments.
Specifically, LMR introduces a context enhancement technique with LLMs to generate crucial target-related context semantics.
These semantics are integrated with visual features for producing discriminative video representations.
Finally, a language-conditioned transformer is designed to decode free-form language queries, on the fly, using aligned video representations for moment retrieval.
Extensive experiments demonstrate that LMR achieves state-of-the-art results, outperforming the nearest competitor by up to 3.28\% and 4.06\% on the challenging QVHighlights and Charades-STA benchmarks, respectively. More importantly, the performance gains are significantly higher for localization of complex queries.

\end{abstract}  
\begin{IEEEkeywords}
Video moment retrieval, multimodal alignment, context enhancement, large language models.
\end{IEEEkeywords}

\section{Introduction}
\label{Intro}
Video Moment Retrieval (VMR) aims to localize the specific moment in an untrimmed video that semantically corresponds to a given language query. 
The success of VMR enhances video analysis with reduced effort and benefits a wide range of applications, including video semantic segmentation, user-friendly video editing, and mass surveillance.
Unlike conventional temporal action localization, constrained by predefined action categories, VMR offers flexibility by localizing complex activities through free-form linguistic expressions.

\begin{figure}
    \centering
\includegraphics[width=\linewidth]{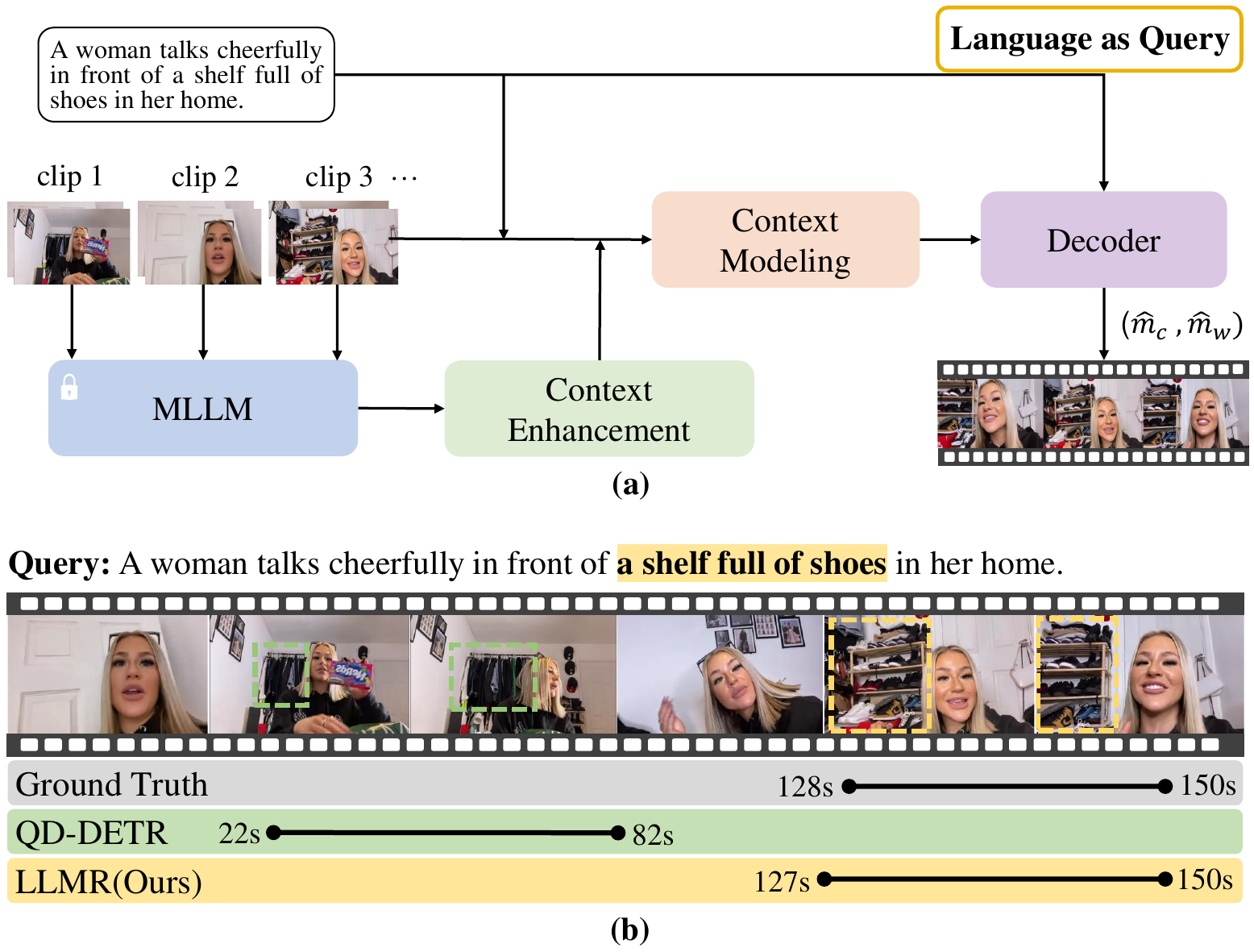}
    \caption{ 
    (a) We introduce LLM-based context enhancement and modeling to improve video representations and directly employ free-form language as queries to decode these representations for accurate moment localization.
    (b) The benchmark QD-DETR~\cite{moon2023query} predicts false positives due to insufficient context modeling and decoding (captures major context but misses important details), whereas, our approach accurately localizes the target moment.
    }
    \label{Intro-case}
\end{figure}


Two popular VMR paradigms are the two-stage matching and coordinate regression. 
The former~\cite{zhang2021multi, chen2020learning, qu2020fine, yuan2019semantic} involves pre-segmenting moment proposals using static multi-scale sliding windows or proposal-generating networks, followed by repeated query-proposal matching to determine the best match. 
The latter paradigm~\cite{moon2023query, liu2022umt, hao2022query, ju2022prompting, chen2021end, zhang2020span} typically uses transformers to directly regress the target start and end coordinates based on the multimodal features of the provided video and sentence query.
Despite promising results, existing methods face challenges in distinguishing the target moment from similar-looking or similar-action ones when query sentences involve detailed environmental, character, and action descriptions. Two-stage matching approaches rely on segment-level features as the proposal representation to estimate proposal-query semantic relevance, often overlooking contextual and global information of videos. For coordinate regression approaches, the transformer's self-attention mechanism aids in distinguishing event boundaries but tends to give high attention values to clips with similar visual appearances, treating them as semantically identical events and thus confusing the model's localization.
As illustrated in \cref{Intro-case}(b), when multiple moments fit the action ``A woman talks cheerfully'' and the background ``a shelf'', the benchmark QD-DETR~\cite{moon2023query} fails to identify the correct moment for ``A woman talks cheerfully in front of a shelf full of shoes''.

Recently, Large Language Models (LLMs) \cite{OpenAI_2023, alayrac2022flamingo, li2023blip, li2023videochat, li2023m} trained with unmatched Internet-scale visual-language data have exhibited impressive capabilities in comprehending visual and textual inputs to promote various vision-language tasks \cite{pan2023retrieving, naeem2023i2mvformer, zang2023contextual, hu2023bliva}.
For instance, contextual object detection models \cite{zang2023contextual} employ the textual output produced by LLMs as prompts to manipulate external vision expert models.
Zero-shot image classification models \cite{naeem2023i2mvformer} use class descriptions generated by LLMs as guidance for classification. 
In VMR, the high-level semantic information contained in LLM-generated text can compensate for visual blind spots and enhance the contextual representation of videos, aiding in distinguishing visually similar events from a textual perspective. However, the potential of LLMs in this role has yet to be explored in VMR.

To this end, we propose Large Language Model-guided Moment Retrieval (LMR), an effective VMR approach that repurposes the inherent knowledge of LLMs to enhance video context modeling and adopts a language-conditioned transformer to decode free-form language queries, on the fly.
Specifically, the proposed LMR generates multiple textual views for each video, including rich contextual details like environmental information, character descriptions, and action narratives.
These crucial views are then distilled into target-related context through Video-Query Fusion (VQF) and combined with visual features to facilitate the alignment between video and intricate language queries and produce highly discriminative video representations.
Finally, we introduce a language-conditioned transformer to directly decode free-form language queries using the highly discriminative video representations for VMR.
Furthermore, considering that queries with longer lengths and more clauses may indicates stricter contextual demands for the queried events, we construct a Complex Query Validation (C-QVal) split from the validation split of an existing dataset based on metrics clause and word counts, and then evaluate the effectiveness of LMR in video retrieval for complex queries.

Our main contributions are summarized as follows.
\begin{itemize}
    

    \item We propose an LLM-guided Moment Retrieval (LMR) method that integrates visual content with target-related context information derived from LLMs for moment retrieval.
    We construct C-QVal dataset with complex queries and show the efficacy of LMR in handling complex scenarios.

    
    \item We devise a language-conditioned transformer to decode free-form language queries, on the fly, using aligned video representations for moment retrieval.


    \item Extensive experiments demonstrate that our approach achieves top-ranked performance on popular benchmarks, outperforming the nearest competitor by up to 3.28\% and 4.06\% on the challenging QVHighlights and Charades-STA, respectively.

\end{itemize}

    




\begin{figure*}[t!]
    \centering
    \includegraphics[width=\linewidth]{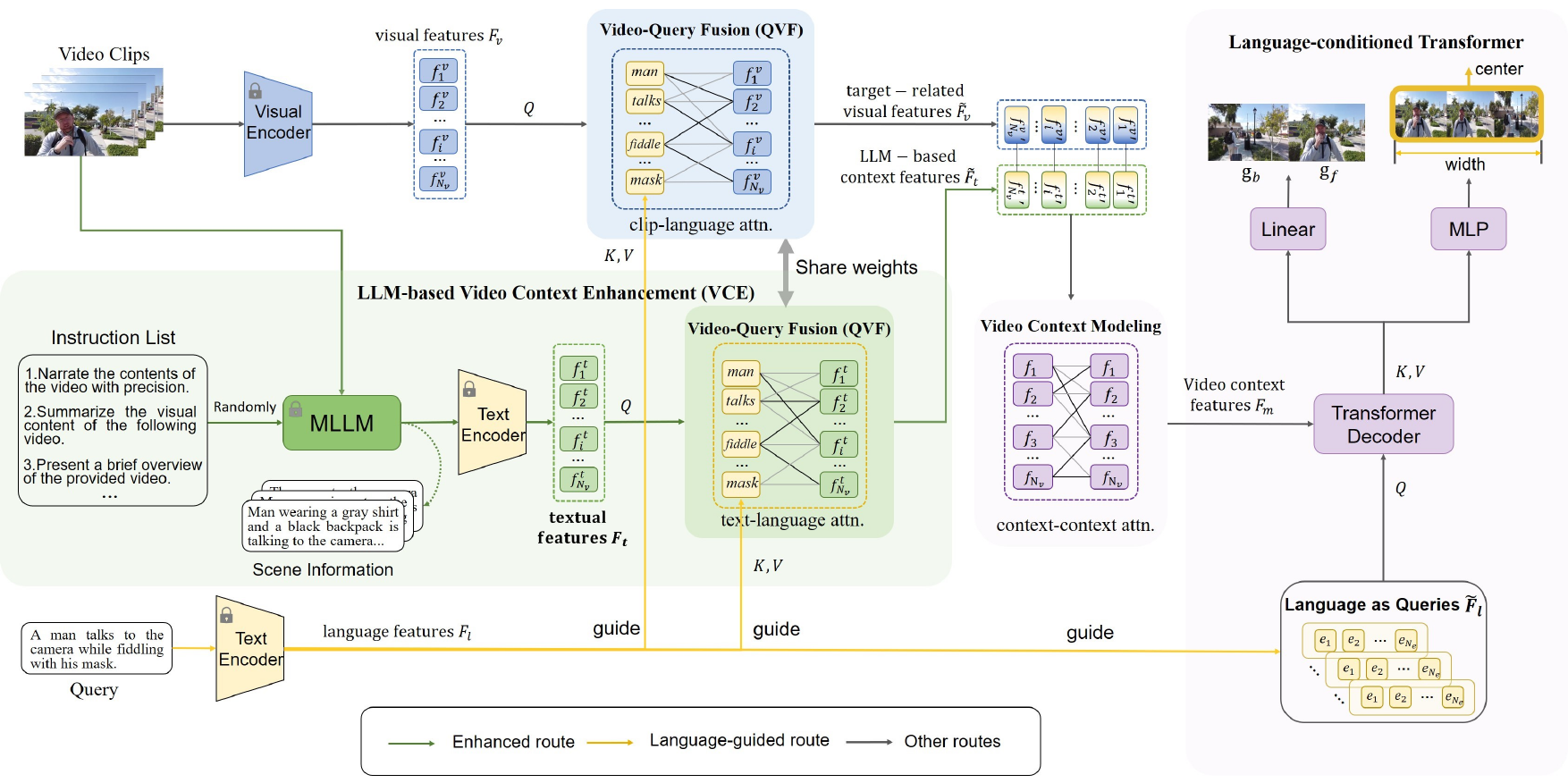}
    \caption{Architecture of LMR. Given a video sequence and a query language, LLM-based Video Context Enhancement generates target-related context information, Video Context Modeling enhances video representations using target-related visual features and LLM-based video context, and Language-conditioned Transformer directly decodes free-form language queries using aligned video representations for moment retrieval.}
    \label{framework}
\end{figure*}

\section{Related Works}
\textbf{Video Moment Retrieval.}
Two popular VMR techniques include the two-stage matching and semantic decomposition paradigms.
The key idea of two-stage matching methods~\cite{zhang2021multi, wang2021structured, chen2020learning, qu2020fine, yuan2019semantic} is to extract a set of moment proposals through multi-scale sliding windows or proposal generating networks, and explicitly model channel-level or sequence-level interactions pairwise for proposal matching and ranking.
Despite their promising performance, generating numerous proposals for different durations and locations to ensure high recall greatly increases computational overhead, negatively impacting the efficiency and difficulty of the subsequent matching process.

Semantic decomposition methods~\cite{li2022compositional, wang2021structured, liu2020jointly, zhang2019man, zhang2019cross, ge2019mac, chen2019semantic} typically decompose video and language into structured hierarchies, such as word-phrase-sentence and event-actions-objects. 
Thus, the semantics of video and query text can be systematically represented by combining different units, and the semantic correlation between text and vision is then learned through interactions among multimodal semantic units. For instance,~\cite{ge2019mac} uses verb-object pairs and activity concept classifiers to encode multimodal features.
Such a decomposition-based paradigm is effective for basic scenarios but struggles to localize moments corresponding to complex situations, where the target moment is conditioned on diverse backgrounds, character appearances, and action sequences.
For such intricate queries, thoroughly representing and understanding the spatiotemporal context in videos is imperative. 
However, existing methods struggle with capturing these detailed semantics.
To address this, we properly employ the extensive knowledge within LLMs to construct target-related video context, thereby enhancing video modeling and representation to handle intricate queries.

\vspace{1mm}
\noindent \textbf{Vision-Language Tasks with LLMs.}
LLMs have showcased remarkable performance in comprehending and generating textual language. They have been successfully applied to a wide range of vision-language tasks, including referring image segmentation~\cite{lai2023lisa, ji2023towards}, image classification~\cite{naeem2023i2mvformer}, contextual object detection~\cite{zang2023contextual}, and video question answering~\cite{yu2024self}. 
These applications leverage the informative textual output produced by LLMs, \eg, ChatGPT\footnote{\url{https://openai.com/blog/chatgpt}}, OPT~\cite{zhang2022opt} and LLaMA~\cite{touvron2023llama}, as prompts to guide external vision expert models.
For instance, contextual object detection models~\cite{zang2023contextual} employ the textual output produced by LLMs as prompts for multimodal context decoding. Zero-shot image classification models~\cite{naeem2023i2mvformer} use class descriptions generated by LLMs as guidance for classification. \cite{yu2024self} utilizes BLIP-2 to generate scores for each frame indicating whether it contains relevant information to answer the question. Despite their success in these fields, the potential of LLMs in VMR has yet to be explored. 

Additionally, the progress in LLMs has also spurred the advancement of multimodal large language models (MLLMs)~\cite{OpenAI_2023, alayrac2022flamingo, li2023blip, li2023videochat, li2023m}. These MLLMs, built on pretrained LLMs, are commonly fine-tuned with extra visual components, have expanded the capabilities of LLMs to comprehend both language and visual inputs. 
Inspired by LLM-based prompting techniques and the success of MLLMs in multimodal tasks, we employ MLLMs to generate multi-view description texts as context prompts for the referred moments.

\section{LLM-guided Moment Retrieval (LMR)}
Given an untrimmed video $V=\{v_i\}_{i=1}^{N^v}$ with ${N^v}$ clips and a query language $Q=\{q_i\}_{i=1}^{N^l}$ with ${N^l}$ words, VMR aims to retrieve the best matching temporal moment $M$ corresponding to the query. 
The target moment $M$ in video $V$ is denoted as 
$M=(m_c, w_\sigma)$, where $m_c$ represents the center coordinate and $w_\sigma$ is the temporal width.

The overall architecture of LMR is shown in ~\cref{framework}. It comprises three main components:
1) LLM-based Video Context Enhancement~(VCE) employs LLM to create target-related textual views based on videos and randomly sampled instructions;
2) Video Context Modeling encodes aligned target-related visual features and multi-view textual context for comprehensive video representations;
3) Language-conditioned Transformer directly uses free-form language as queries to identify referred moments based on these video representations.

\subsection{Visual and Textual Encoders}  
\label{Feature Encoding}
Following the  previous standard~\cite{lei2021detecting,liu2022umt,moon2023query}, we use CLIP~\cite{radford2021learning} and SlowFast~\cite{feichtenhofer2019slowfast} as visual backbones to extract visual features $F^v =\{f^v_i\}_{i=1}^{N^v}$ and CLIP-Text as the textual backbone to extract language features $F^l =\{f^l_i\}_{i=1}^{N^l}$,
where $f^v_i$ and $f^l_i$ denote the visual features of the $i$-th video clip and the language features of the $i$-th word, respectively.

\subsection{LLM-based Visual Context Enhancement}\label{VCE}
Although MLLMs can handle both video and text inputs simultaneously, their direct application to the VMR task is unreliable.
This is because MLLMs are primarily optimized for knowledge-grounded text generation, without explicit focus on temporal reasoning in their fine-tuning objectives.
As a result, we properly leverage the impressive text generation capabilities of MLLMs (with frozen weights) to produce textual descriptions for video clips. 
By extracting target-related context information from the generated descriptions, \eg, the background of the current clip and the appearance of individuals, we strengthen the contextual semantic representations of videos, enabling the model to distinguish moments with similar appearances or similar actions.


\noindent \textbf{Video Context Generation.}  
In this work, we employ VideoChat~\cite{li2023videochat}, which is built upon BLIP-2~\cite{li2023blip} and StableVicuna\footnote{\url{https://github.com/Stability-AI/StableLM}}, as the context generator given their exceptional ability in video comprehension. 
Specifically, we generate an instruction list using ChatGPT for VideoChat, with an example instruction referencing LLaVA\cite{liu2024visual}. The complete instruction list is provided in \cref{instruction_list}. 
For each clip of a video, the instruction used in VideoChat is randomly selected from the list, enabling it to produce diverse descriptions that accommodate query texts with ever-shifting styles, from detailed descriptions like ``A woman in an orange blouse conversing with a man in a light blue shirt in front of the camera'' to brief ones like ``person talking on the phone''. After generating descriptions and organizing them in chronological order, a temporal descriptive context is established for the video sequence. 
We finally extract target-related information from the temporal descriptions to augment the video representations for accurate localization.

\begin{table}[t]
\renewcommand{\arraystretch}{1}
    \centering
    \caption{List of instructions used for VideoChat~\cite{li2023videochat} in video description generation.}
    \resizebox{0.49\textwidth}{!}{%
    \begin{tabular}{l}
    \toprule
         1.\quad "Describe the following video concisely." \\
         2.\quad"Present a brief overview of the provided video." \\
         3.\quad"Provide a concise description of the given video." \\
         4.\quad"Convey a short narrative summarizing the provided  video." \\
         5.\quad"Summarize the visual content of the following video." \\
         6.\quad"Deliver a compact portrayal of the presented video." \\
         7.\quad"Furnish a concise explanation of the given video." \\
         8.\quad"Supply a brief account of the provided video." \\
         9.\quad"Narrate the contents of the video with precision." \\
         10.\quad"Offer a succinct analysis of the given video." \\
    \bottomrule
    \end{tabular}
    }
    \label{instruction_list}
\end{table}

Let us denote language tokens as $o^l\in \mathbb{R}^{d_1\times N^s}$, where $d_1$ and $N^s$ represent the input dimension of LLM and the length of instruction language.
After applying a projection layer, the visual features $f^v_i$
are transformed into corresponding visual tokens $o^v\in \mathbb{R}^{d_1\times N^c}$, where $N^c$ represents the length of visual tokens for a clip. 
LLM is able to decode multimodal contextual tokens conditioned on a prefix $\textbf{e}_{1:n}$. Here the prefix is the concatenation of the projected visual tokens $o^v$ and the language tokens as $o^l$, \ie, $\textbf{\textit{e}}_{1:n}=[o^v,o^l] \in \mathbb{R}^{d_1\times(N^c+N^s)}$, acting as the generated target in an autoregressive way:
\begin{equation}
    \textit{p}(\textbf{\textit{e}}_{n+1:N^c} | \textbf{\textit{e}}_{1:n}) =  
    \prod^{N^c}_{i=n+1}{p_\theta(\textbf{e}_{i+1} | \textbf{e}_{1:i} )}.
\end{equation}

To generate new tokens, the LLM first predicts the latent embedding $\textbf{\textit{e}}_{n+1}$ for the next $n+1$-th token through multiple Transformer Layers (denoted as $\phi$), and then uses a Feed Forward Network (denoted as $\mathcal{F}$) to compute the probability distribution $\textit{p}(\textbf{\textit{h}})$  
\begin{gather}
     \textbf{\textit{h}}_{n+1} = \phi(\textbf{\textit{e}}_{1:n}), \\
     \textit{p}(\textbf{\textit{e}}_{n+1})={\rm softmax}(\mathcal{F}(\textbf{\textit{h}}_{n+1})),
\end{gather}
where the token $\textbf{\textit{e}}_{n+1}$ is an element from the vocabulary, representing human words in natural language. The autoregressive generation process continues until the language token $\textbf{\textit{e}}_{N^c}$ hits the [EOS] token, signifying the completion of the description text $t_i$ corresponding to the video clip $v_i$. We simultaneously perform description generation for all clips of an input video $V$, and stack them sequentially to obtain the temporal context description $T = \{t_i\}^{N^v}_{i=1}$.

\vspace{1mm}
\noindent\textbf{Video-Query Fusion (VQF).} Considering that LLM-generated description text contains redundant semantic information between video clips, we construct cross-attention layers to highlight referred moment-related context information based on query language. 
Specifically, initial textual representation $F^t =\{f^t_i\}_{i=1}^{N^t}$ and query representation $F^l =\{f^l_i\}_{i=1}^{N^l}$ are extracted by a frozen text encoder, where $f^t_i$ represents the textual feature of $i$-th video clip and $N^t$ is the number of description texts.
For better feature interaction in cross-attention layers, we keep the language feature $F^l$ and the textual feature $F^t$ in a unified semantic space by extracting both with the same text encoder. 

In each cross-attention layer, $F^t$ serve as \textit{query} and language features $F^l$ provide \textit{key} and \textit{value}:
\begin{equation}
    {\rm Attn}(Q^v, K^l, V^l) = {\rm softmax}(\frac{Q^v (K^l)^T}{\sqrt{d_1}})V^l
\end{equation}
Here, $K^l, V^l, Q^v$ are projected language and textual feature and $d_1$ representes their feature dimension.
Then the textual feature $\tilde{F}^{t}$ can be represented as weighted sum of $F^l$. Following the standard transformer architecture, the attention scores are projected through a Multilayer Perceptron (MLP) and incorporated into the original textual representations. We refer the output of this module as LLM-based context feature, denoted as $\tilde{{F}^t} =\{{f^t}'_i\}_{i=1}^{N^t}$.


\subsection{Video Context Modeling}
Video Context Modeling enhances video representations with visual features and LLM-based video context, taking inputs from the LLM-based context feature $\tilde{F}^t$ and the target-related visual feature $\tilde{F}^v$. 
Here, $\tilde{F}^v$ is also computed with a VQF module, similarly to $\tilde{F}^t$ in \cref{VCE}. Due to the temporal consistency between LLM-generated context $F_t$ and visual features $F_v$, the two VQF modules share weights, facilitating joint learning of query relevance in their respective contexts. 
The target-related visual features $\tilde{F}^v$ and LLM-based context $\tilde{F}^t$ are then concatenated to form the video contextual sequence. Additionally, a randomly initialized saliency token is prepended to the sequence to predict the relevance score between video and text query. This sequence is encoded through two self-attention layers, yielding LLM-enhanced video representations $F_m$.

The video tokens $\{x_i\}_{i=1}^{N^v}$ within $F_m$ serve for both decoding and video-query relevance computation. Decoding details are described in \cref{decoding}. For the latter, the saliency token $x_s$ and video tokens are projected using a single fully-connected layer, $w_s$ and $w_v$, respectively. Thus, the predicted relevance score $S_i$ between clip $v_i$ and text query $Q$ is computed as:
\begin{equation}
    S_i = \frac{w^T_s x_s \cdot w^T_v x_v^i}{\sqrt{d}}
\end{equation}
where $d$ is the channel dimension of projected tokens. Subsequently, $S_i$ and $\{x_i\}_{i=1}^{N^v}$ undergo contrastive learning (in \cref{loss}) and decoding, respectively.

\subsection{Language-conditioned Transformer} \label{decoding}
The DETR-based encoding-decoding architecture is widely adopted in detection and retrieval tasks~\cite{lei2021detecting, moon2023query}. It originates from the traditional object detection tasks, aiming to predict a set of bounding boxes and category labels for objects of interest when the exact search target is unknown. 
However, different from object detection, moment retrieval precisely knows the search target, \ie, the video segment corresponding to the query language. Therefore, using randomly initialized query embeddings as the input for the decoder is not suitable for known-target moment retrieval.
However, most existing methods overlook this distinction and apply DETR in a straightforward manner. Randomly initialized embeddings ($\textit{query}$) struggle to effectively capture referred event related feature information from the visual sequence ($\textit{key,value}$) in the cross-attention calculation at the decoder layer.

In our work, we address this issue by employing a small set of moment queries, as the input for the decoder. Consequently, all the queries are obligated to find the referred moment only.
Specifically, we replicate the query language features $F^l$ $k$ times to create decoder embedding, \ie moment queries, together with trainable positional parameters as the position encoding, to encode temporal information, where $k$ is a hyperparameter. 


Note that the query language has been fused into both target-related visual features and LLM-based context features before being passed into the VCM module. The video context features output by VCM serve as the \textit{K,V} in the decoder. Therefore, during the decoding process, the early-aligned video representations and the query language can better learn their co-relationship using the attention mechanism.

As a result, all moment queries are constrained by the language expressions, ensuring that they target only the referred moment, even when similar events exist in the video. One query will emerge with a significantly higher score, while other query scores will be suppressed. We define the output of the Transformer decoder as $H=\{h_i\}_{i=1}^{N^v+N^t}$, $h_i \in \mathbb{R}^{d_1}$.

\subsection{Localization and Loss Functions} \label{loss}
Finally, the temporal bounding box of the target moment is predicted based on the language-conditioned visual features $H$. As shown in ~\cref{framework}, we feed the language-conditioned visual features into a MLP layer to compute the center coordinate $\hat{m}_c\in [1,N^v]$ and window width $\hat{m}_w$ of the predicted moment $\hat{y}$, and a linear layer to compute the class $g$ either foreground $g_f$ or background $g_b$. 

The loss $\mathcal{L}_{mr}$ is used to measure the discrepancy between the prediction and ground truth moments. Following~\cite{rezatofighi2019generalized, moon2023query, lei2021detecting}, it consists of three losses: a $\mathcal{L}1$ loss for $m_c$, a generalized IoU loss $\mathcal{L}_{IoU}$ for $m_c, m_w$  and a cross-entropy loss $\mathcal{L}_{CE}$ for $g$. Therefore, $\mathcal{L}{mr}$ is calculated as follows:
\begin{small}
    \begin{gather}
        \mathcal{L}_{mr} = \lambda_{\mathcal{L}1}||m_c-\hat{m}_c|| + \lambda_{IoU}\mathcal{L}_{IoU}(m_w,\hat{m}_w) + \lambda_{CE}\mathcal{L}_{CE}, \\
        \mathcal{L}_{CE}=-\sum_{y\in \mathcal{G}}{y}log(\hat{y}), \mathcal{G}=[g_f,g_b], 
\end{gather}
\end{small}
\noindent where $(m_c,m_w)$ are the center coordinates and window width of the ground truth moments $y$, and $\lambda_*\in \mathbb{R}$ are hyperparameters balancing the terms. 
Moreover, we treat sentences from other samples within batches as negative samples, which should have lower relevancy scores $S$ to the videos, and introduce inter-video contrastive learning for cross-sample supervision:
\begin{equation}
    \mathcal{L}_{cont} = -log(1-S_{neg})
\end{equation}
Finally, our overall training objective is defined as:
\begin{equation}
    \mathcal{L} = \lambda_{mr}\mathcal{L}_{cont} + \lambda_{cont} \mathcal{L}_{mr}
\end{equation}




\section{Experiments}
\subsection{Datasets and Evaluation Metrics}
In order to validate the effectiveness of the proposed LMR, we conducted extensive experiments on two publicly available datasets: Charades-STA~\cite{sigurdsson2016hollywood} and QVHighlights~\cite{lei2021detecting}.

\vspace{1mm}
\noindent \textbf{Charades-STA}~\cite{sigurdsson2016hollywood} is a dataset for temporal sentence grounding. Derived from Charades, it provides 12,408 training and 3,720 testing moment-sentence pairs. The videos capture human actions and queries, offering an average video length of 29.8 seconds. 

\vspace{1mm}
\noindent \textbf{QVHighlights}~\cite{lei2021detecting} is a dataset for moment retrieval and highlight detection, offering over 10,000 videos of 150-seconds each, along with human-written text queries. These queries describe relevant video segments averaging 24.6 seconds. The dataset contains 10,310 queries and 18,367 moments. We adopt the original data splits, using the train split for model training and the test split for evaluation.

\vspace{1mm}
\noindent \textbf{Evaluation Metrics.}
We use evaluation metrics 'R@n, IoU=m'~\cite{li2023g2l,li2023d3g,yan2023unloc,lin2023univtg} which calculate the percentage of language queries having at least one correct retrieval (temporal IoU with ground truth moment is larger than m) in the top-n retrieved moments. Following standard practice, we use $n \in {1, 5}$ and $m \in {0.5, 0.7}$. For the Charades-STA dataset, we also adopt the mean average precision (mAP) at various thresholds, similar to prior works~\cite{liu2022umt,lei2021detecting}.

\begin{table*}
\centering
\renewcommand{\arraystretch}{1.15}
    \addtolength{\tabcolsep}{0pt}
    \caption{Performance comparison on QVHighlights \textit{test} split. All methods use only videos (and no audio) as data source and Slowfast and CLIP as the visual backbones.}
    \resizebox{0.7\textwidth}{!}{%
    \begin{tabular}{lcccccc}
    \hline
          Model & \quad R1@0.5\quad & \quad R1@0.7\quad & mAP@0.5 & mAP@0.75 & mAP Avg. \\
    \hline\hline
         MCN~\cite{anne2017localizing} &  11.41 & 2.72 & 24.94 & 8.22 & 10.67\\
         CAL~\cite{gao2017tall} &  25.49 & 11.54 & 23.40 & 7.65 & 9.89\\
         XML~\cite{lei2020tvr} &  41.83 & 30.35 & 44.63 & 31.73 & 32.14\\
         XML+~\cite{lei2020tvr} &  46.69 & 33.46 & 47.89 & 34.67 & 34.90\\
         M-DETR~\cite{lei2021detecting} &  52.89 & 33.02 & 54.82 & 29.40 & 30.73\\
         UniVTG~\cite{lin2023univtg} &  58.86 & 40.86 & 57.60 & 35.59 & 35.47\\ 
         UMT$^\dagger$~\cite{liu2022umt} &  56.23 & 41.18 & 53.38 & 37.01 & 36.12\\
         QD-DETR~\cite{moon2023query} &  62.40 & 44.98 & 62.52 & 39.88 & 39.86\\
    \hline      
         \textbf{Ours}  & \textbf{64.40} & \textbf{47.21} & \textbf{64.65} & \textbf{43.16} & \textbf{42.56}\\
    \bottomrule
    \end{tabular}
    }
    \label{QVHighlights}
\end{table*}

\subsection{Implementation Details}
For a fair comparison, we simply leverage the pre-extracted SlowFast~\cite{feichtenhofer2019slowfast} and CLIP~\cite{radford2021learning} features on QVHighlights and official VGG~\cite{simonyan2014very} features on Charades-STA, provided by other baselines: Moment-DETR~\cite{lei2021detecting} and UMT~\cite{liu2022umt}, respectively.  For query text and generated description text, we use only the frozen embedding layer of Llama~\cite{touvron2023llama}. On QVHighlights, we use the query embeddings from CLIP instead before performing cross-attention with visual features, bringing them into the same feature spaces. The encoder is composed of 4 layers of transformer block (2 cross-attention layers and 2 self-attention layers) while the decoder has two layers, following~\cite{moon2023query}. 
For LLM, we employ the VideoChat~\cite{li2023videochat}, which is built on BLIP-2~\cite{li2023blip} and StableVicuna\footnote{\url{https://github.com/stability-AI/}} and remains frozen throughout its usage. The weights of losses are set as $\lambda_{mr}=1$, $\lambda_{cont}=1$, $\lambda_{\mathcal{L}_1}=10$, $\lambda_{\mathcal{L}_{IoU}}=1$ and $\lambda_{\mathcal{L}_{CE}}=4$. The hidden Sdimensions of transformers are set to 256, with no expansions in feed-forward networks. In all experiments, we use Adam~\cite{kingma2014adam} optimizer with 1e-4 learning rate and 1e-4 weight decay. The model is trained with batch size 32 for 200 epochs on QVHighlights, and batch size 8 for 100 epochs on Charades-STA. 
All experiments are conducted on a single RTX 3090 GPU. LLM is utilized offline only once for video preprocessing, and our model's training process requires only 2-5 hours. Furthermore, inference on a video takes only 0.45 seconds after training.

\subsection{Comparisons with State-of-the-art Methods}
\textbf{Comparison on QVHighlights.} In \cref{QVHighlights}, we report our performance compard to other state-of-the-art (SOTA) methods on QVHighlights dataset. For a fair comparison, all methods in \cref{QVHighlights} only use video as the data source and no audio. 
As observed, LMR stands out by achieving substantial improvements, especially when evaluated with stricter IoU$>$0.5. It surpasses previous SOTA methods by significant margins, with gains of up to 2.23\% and 3.28\% on R1@0.7 and mAP@0.75, respectively. It is worth noting that even when QD\_DETR uses both video and audio as data sources, LMR still outperforms it across all evaluation metrics while relying solely on video source. Specifically, LMR exceeds QD\_DETR by 1.34\%, 2.11\%, 1.61\%, 3.06\% on R1@0.5, R1@0.7 mAP@0.5, mAP@0.75 and mAP@Avg, respectively. These results verify the effectiveness of employing LLMs to model multimodal contexts in the domain of video moment retrieval.

\begin{table*}
\centering
\renewcommand{\arraystretch}{1.15}
    \setlength{\tabcolsep}{5.5pt}
    \caption{Performance comparisons on Charades-STA \textit{test} split. All methods use videos as data sources and VGG features.}
    \resizebox{0.7\textwidth}{!}{%
     \begin{tabular}{lcccc}
     \toprule
         Method & R@1, IoU=0.5 & R@1,IoU=0.7 & R@5,IoU=0.5 & R@5,IoU=0.7 \\        
    \hline\hline
            MCN~\cite{anne2017localizing} & 27.42 & 13.36 & 66.37 & 38.15 \\
            SAP~\cite{chen2019semantic} & 27.42 & 13.36 & 66.37 & 38.15 \\
            MAN ~\cite{zhang2019man} & 41.21 & 20.54 & 83.21 & 51.85 \\
            2D-TAN~\cite{zhang2020learning} & 40.94 & 22.85 & 83.84 & 50.35 \\
            FVMR~\cite{gao2021fast} & 42.36 & 24.14 & 83.97 & 50.15 \\
            CBLN~\cite{liu2021context} & 43.67 & 24.44 & 88.39 & 56.49 \\
            MMRG~\cite{zeng2021multi} & 44.25 & - & 60.22 & - \\
            MMN~\cite{wang2022negative} & 47.31 & 27.28 & 83.74 & 58.41 \\
            G2L~\cite{li2023g2l} & 47.91 & 28.42 & 84.80 & \textbf{59.33} \\
            UMT~\cite{liu2022umt} & 48.31 & 29.25 & \textbf{88.79} & 56.08 \\
            QD-DETR~\cite{moon2023query} & 52.77 & 31.13 & -  \\
        \hline
            \textbf{Ours} & \textbf{55.91} & \textbf{35.19} & 83.79 & 50.48 \\  
    \bottomrule
    \end{tabular}
    }
    \label{Charades}
    \vspace{5pt}
\end{table*}

\noindent \textbf{Comparison on Charades-STA.} \cref{Charades} presents a detailed performance comparison with other state-of-the-art (SOTA) methods on the Charades-STA dataset. In \cref{QVHighlights}, we emphasize again that all methods use only video as the sole data source without using audio for a fair comparison. The results clearly demonstrate the proposed LMR's significant enhancements on R@1 metrics. LMR outperforms previous SOTA methods by substantial margins, achieving impressive improvements of 3.14\% and 4.06\% on the R1@0.5 and R1@0.7 metrics, respectively. Remarkably, even when QD\_DETR employs both video and audio as data sources, LMR maintains its superior performance across R1@* evaluation metrics when relying solely on video source. More specifically, when assessed with VGG features, LMR surpasses QD\_DETR by 0.4\% and 1.02\% on the R1@0.5 and R1@0.7 metrics, respectively. These results underscore the effectiveness of integrating knowledge of LLMs for multimodal contextual semantic modeling. Furthermore, it is important to note that R1@* metrics are more challenging than R5@*, requiring models to achieve high levels of semantic alignment accuracy by satisfying IoU requirements with only one predicted moment. These results verify the significance of augmented query text in guiding the decoding process to focus on the most semantically relevant moments.

\begin{figure*}
    \centering
    \includegraphics[width=0.7\linewidth]{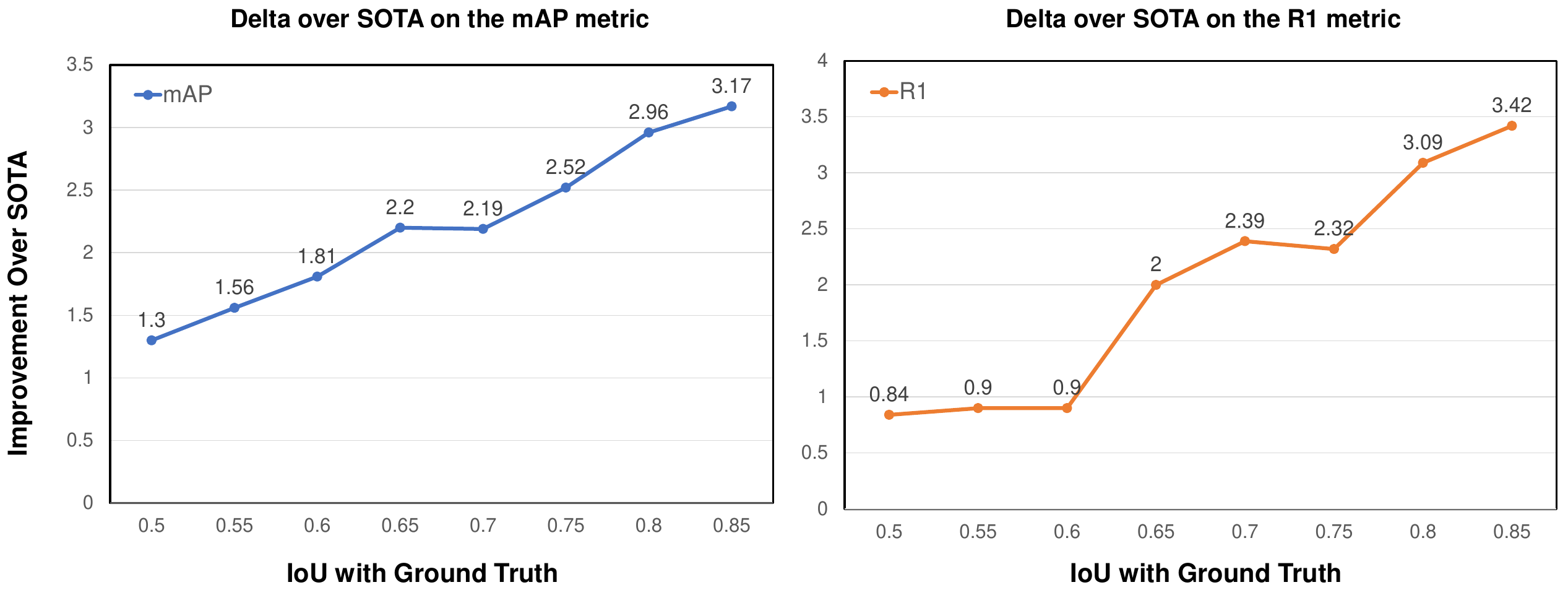}
    \caption{Improvements over the state-of-the-art (QD-DETR \cite{moon2023query}) on mAP and R1 metrics on the Charades dataset.}
    \label{delta over SOTA}
\end{figure*}

Furthermore, we investigate the performance improvements of our LMR compared to the previous state-of-the-art (QD-DETR\cite{moon2023query}) across dense IoU values on mAP and R1 metrics. In \cref{delta over SOTA}, the horizontal axis represents IoU values in increments of 0.05, while the vertical axis indicates the percentage of performance improvement of LMR over the state-of-the-art. It can be observed that as the IoU requirements with ground truth increase, the relative improvement of LMR over state-of-the-art also gradually increase. This indicates that LMR's superiority becomes more pronounced when stricter requirements for moment localization are imposed. These results verify the significant enhancement of the proposed LMR in semantic alignment between language and vision as well.

\begin{table*}[t]
\renewcommand{\arraystretch}{1.4}
    \centering
    \begin{threeparttable}
    \caption{Comparison to the SOTA method QD-DETR\cite{moon2023query} on the localization of complex queries on multiple \textit{C-QVal} split.}
    \setlength{\tabcolsep}{4.2pt}
    \small
    \begin{tabular}{p{1cm}<{\centering}ccccccccccccccc}
    \toprule
        \multirow{2}{*}{Metrics} & \multirow{2}{*}{Model} & \multicolumn{13}{c}{\textbf{(Least Clause Count, Least Word Count)}} \\
    \cline{3-15}
    \renewcommand{\baselinestretch}{1.2}
        ~  & ~ & (1,8) & (1,9) & (1,10) & (1,11) & (1,12) & (1,13) & (1,14) & (1,15) & (1,16) & (2,8) & (2,9) & (2,10) & \textbf{Avg} \\
    \hline\hline
        \multirow{3}{*}{R1@0.7}~&~QD\_DETR~&~46.39~&~45.89~&~44.59~&~43.61~&~44.69~&~42.55~&~44.44~&~44.00~&~37.93~&~50.00~&~51.02~&~50.00&45.43\\
         ~&~LMR~&~54.12~&~52.97~&~52.46~&~53.01~&~53.98~&~53.72~&~52.94~&~52.8~&~51.72~&~54.00~&~55.10~&~54.17~&53.42\\
         \rowcolor[gray]{0.9} 
        ~ & \textbf{Perf. Gain} & \textbf{+7.73} & \textbf{+7.08} & \textbf{+7.87} & \textbf{+9.40} & \textbf{+9.29} &\textbf{+11.17} &\textbf{+8.50} & \textbf{+8.80} & \textbf{+13.79} & \textbf{+4.00} &\textbf{+4.08} & \textbf{+4.17}&\textbf{+7.99}\\
    \hline
        \multirow{3}{*}{mAP@0.7}~&~QD\_DETR~&~43.96~&~43.54~&~41.94~&~39.92~&~41.06~&~40.72~&~42.68~&~40.95~&~34.86~&~45.24~&~46.17~&~46.09&42.26\\
        ~&~LMR~&~50.15~&~49.01~&~48.58~&~47.25~&~48.50~&~49.08~&~49.58~&~48.91~&~45.16~&~52.95~&~54.03~&~54.12~&49.78\\
        
        \rowcolor[gray]{0.9} ~&\textbf{Perf. Gain}&\textbf{+6.19}&\textbf{+5.47}&\textbf{+6.64}&\textbf{+7.33}&\textbf{+7.44}&\textbf{+8.36}&\textbf{+6.90}&\textbf{+7.96}&\textbf{+10.30}&\textbf{+7.71}&\textbf{+7.86}&\textbf{+8.03}&\textbf{+7.52}\\
    \bottomrule
    \end{tabular}
        \begin{small}
            \begin{tablenotes}[flushleft]
              \item (clause count, word count) represents the minimum number of clauses and words simultaneously included in a query. Higher values indicate greater complexity of the queries, implying more stringent contextual requirements for locating the activity moments.
            \end{tablenotes}
        \end{small}
    \end{threeparttable}
    \label{exp on complex query}
\end{table*}

\subsection{Comparison on Complex Query split} 
\textbf{Construction of \textit{Complex Query} (C-QVal) split.} To evaluate the effectiveness of our proposed LMR in video retrieval for complex queries, we extract splits of complex queries from the val split of the QVHighlights dataset, referred to as ``\textit{C-QVal}'' split. Considering that queries with longer lengths and more clauses may have a higher chance of including strict contextual requirements, such as video background requirements, character appearance specifications, and action constraints, etc., we determine whether a query belongs to complex queries based on the presence of clauses and word numbers. Moreover, the complexity of a query increases with the simultaneous inclusion of more clauses and words, indicating higher contextual demands for the queried activities and greater localization challenges. Specifically, we utilize the syntactic analysis library spaCy\footnote{\url{https://spacy.io/}} from the NLP field to calculate the number of clauses in a query. We set different thresholds for the number of clauses and words to obtain \textit{C-QVal} split with varying levels of complexity. For complex query extraction under different parameter settings~(clauses and word count), we obtained multiple sets of \textit{C-QVal} splits with varying numbers of queries, ranging from 48 to 388.
After training the model on the complete \textit{train} split of the QVHighlights dataset, we perform inference on \textit{C-QVal} split and compute metrics to validate the model's alignment capability for complex queries.

\noindent \textbf{Comparison with SOTA Method.} \cref{exp on complex query} presents detailed comparison results with the SOTA method QD-DETR~\cite{moon2023query} on multiple \textit{C-QVal} splits. Compared with QD-DETR, our method demonstrates a significant performance improvement across all \textit{C-QVal} splits, achieving an average boost of 7.99\% and 7.52\% on the most challenging evaluation metrics R@1 IoU=0.7 and mAP IoU=0.7, respectively. This validates the effectiveness of LMR in complex query localization. Moreover, the boosts for both metrics generally exhibit an increasing trend with the growing difficulty of \textit{C-QVal} splits, reaching their peaks at (least clause count, least word count)=(1,16) with boosts of up to 13.79\% and 10.30\%, respectively. This suggests the tremendous potential of LMR in localizing complex contextualized queries. The experimental results above confirm that fusing visual content with complementary context information derived from LLMs has a profound impact on enhancing the semantic alignment between videos and complex contextualized queries.

\begin{table}[t!]
\renewcommand{\arraystretch}{1.2}
    \centering
    \begin{threeparttable}
        \caption{Ablation on LMR components.}
        \begin{tabular}{l|ccccc}
        \toprule
             \multirow{2}{*}{Model} & R1 & R1 & mAP & mAP & mAP \\
              ~ & @0.5 & @0.7 & @0.5 & @0.75 & Avg. \\
        \hline\hline             
             w/o VCE & 60.55 & 44.39 & 60.18 & 39.77 & 38.68\\
             w/o VFE & 55.13 & 37.81 & 56.51 & 34.83 & 33.14\\ 
             w/o LMQ & 63.87 & 47.39 & 63.36 & 42.49 & 42.12\\
             Full Model & 65.48 & 48.58 & 65.06 & 44.45 & 43.46 \\
        \bottomrule
        \end{tabular}
        \begin{small}
        \begin{tablenotes}[flushleft]      
          \item VCE, VFE, and LMQ denote LLM-based Video Context Enhancement, Visual Feature Extraction, and Language-conditioned Moment Queries.
        \end{tablenotes}
        \end{small}
    \end{threeparttable}
    \label{ablation_LMR}
\end{table}


\subsection{Ablation Study}
Here, we investigate the effectiveness of the individual components within the proposed LMR as well as the effectiveness of the internal design in LLM-based Video Context Enhancement module~(VCE). These experiments are conducted on the QVHighlights validation set.

\noindent \textbf{Effectiveness of Individual Components.}
we conduct a thorough ablation study on the three main components in LMR to verify their effectiveness. As depicted in \cref{ablation_LMR}, the removal of the entire LLM-based Video Context Enhancement (w/o VCE) leads to a performance degradation of up to 4.54\%, underscoring the significance of incorporating contextual scene information from LLMs. Additionally, from w/o VFE, we observe that the model can still function normally without the original video-based visual features, although there is obvious  performance drop. The combined results from w/o VCE and w/o VFE demonstrate that visual features play an important role in VMR. However, LLM-generated temporal scene information effectively provides complementary contextual details, enhancing localization performance. In experiments without LMQ, we replace Language-conditioned positional embedding with randomly initialized positional embedding as the decoder input. This leads to an average performance drop of 1.56 points, emphasizing the importance of using language as moment queries for improved decoder guidance in identifying the target moment.

\begin{table}[t!]
\renewcommand{\arraystretch}{1.2}
    \centering
    \resizebox{0.48\textwidth}{!}{%
    \begin{threeparttable}
        \caption{Ablation of components within VCE.}
        \begin{tabular}{l|ccccc}
        \toprule
             \multirow{2}{*}{Model} & R1 & R1 & mAP & mAP & mAP \\
              ~ & @0.5 & @0.7 & @0.5 & @0.75 & Avg. \\
        \hline\hline             
             repl. $LLM_{LLaVA}$  & 61.61 & 46.95 & 62.60 & 43.41 & 42.13\\
             repl. $LLM_{ImgCap}$ & 60.82 & 44.98 & 61.76 & 41.03 & 39.91\\  
             w/o WS & 63.35 & 48.13 & 63.59 & 44.36 & 43.12\\  
             w/o VQF & 63.22 & 46.50 & 62.87 & 42.48 & 41.30\\   
             Full Model & 65.48 & 48.58 & 65.06 & 44.45 & 43.46 \\
        \bottomrule        
        \end{tabular}
        \begin{tablenotes}[flushleft]
            \item repl. LLM (LLaVA): replace the video-based LLM VideoChat with a image-based LLM LLaVA~\cite{liu2024visual}; repl. LLM (ImgCap): replace LLM with a general image captioning model~\cite{anderson2018bottom}; w/o WS: no weight sharing between the two VQFs;  w/o VQF: remove VQF.
        \end{tablenotes}
    \end{threeparttable}
    \label{ablation_VCE}
    }
    \vspace{-14pt}
\end{table}


\noindent \textbf{Effectiveness of LLM-based Video Context Enhancement.}
In \cref{ablation_VCE}, we conduct a detailed ablation study on the internal design of VCE. 
For the LLM replacements (LLaVA and ImgCap), we replace the video-based LLM VideoChat~\cite{li2023videochat} with an image-based LLM LLaVA-1.6~\cite{liu2024visual} and a general image captioning model UTD~\cite{anderson2018bottom} to generate image captions. The images are sampled every 2 seconds in videos, and we apply the same instructions for LLaVA as those used in VideoChat. As illustrated in \cref{ablation_VCE}, the performance of LLaVA replacement exhibits a decrease of 1.04\% to 3.87\%, while ImgCap experiences a more significant decline, ranging from 3.3\% to 4.66\%. This is because, even though image-based LLMs lack temporal and audio information like video-based LLMs, their generated captions can still offer more valid and detailed scene information than general image captioning models, given that LLMs are pretrained on web-scale data.
In w/o WS, the model shows a significant 2.13\% drop in R1@0.5, verifying the advantages of weight sharing between VQFs.  This is due to the temporal correspondence between LLM-generated scene information and visual features, which enables joint learning of query relevance in their respective contexts. 
Directly removing VQF in VCE results in a performance degradation of 2.13\% on average, highlighting the effectiveness of using query language to emphasize related context information derive from LLM.

\textbf{Ablation Studies on Language-conditional Moment Query Number}.
In \cref{ablation-query_number}, we evaluate the performance of our model with different numbers of Moment Queries. 
Previous methods~\cite{lei2021detecting, moon2023query} need to carefully set the number $k$ of Moment Queries to achieve good performance.
For example, reducing $k$ from 20 to 10 in~\cite{lei2021detecting} leads to substantial drops in R1@0.5, R1@0.7, and average mAP metrics by 6\%, 5.74\%, and 7.39\%, respectively.
In contrast, our model consistently achieves good performance for different values of $k$, even with $k=1$. 
This benefits from the design of language-conditioned moment queries focusing on localizing referred activities.
Moreover, our model shows potential for further improvement since its accuracy increases with the number of moment queries. 
For instance, increasing $k$ from 10 (adopted in our main experiments to align with prior works) to 20 boosts the R1@0.7 score of our model by 1.5\%.
We conjecture that since language-conditional moment queries provide explicit retrieval targets during decoding, additional moment queries enable the model to generate a larger number of quality moment candidates, enhancing its ability to handle complex video scenes with multiple similar activities. 


\begin{table}[h]
    \caption{Ablation study for the number ($k$) of language-conditional moment queries on QVHighlights validation split.}
    \centering
    \resizebox{0.49\textwidth}{!}{%
    \begin{tabular}{c|ccccc}
    \toprule
         k & \quad R1@0.5\quad & \quad R1@0.7\quad & mAP@0.5 & mAP@0.75 & mAP Avg. \\
    \hline\hline
         1  & 62.84 & 48.58 & 59.10 & 41.54 & 40.00\\
         3  & 62.84 & 48.58 & 59.10 & 41.54 & 40.08 \\
         5  & 63.03 & 48.19 & 60.91 & 41.15 & 40.60 \\ 
         8  & 63.87 & 48.13 & 63.79 & 42.8 & 42.17\\  
         10 & 63.74 & 49.16 & 63.96 & 44.03 & 43.45\\
         20 & 64.26 & 50.65 & 64.73 & 45.37 & 44.52 \\
         30 & 64.65 & 50.06 & 64.64 & 45.26 & 44.00 \\
    \bottomrule
    \end{tabular}
    }
    \label{ablation-query_number}
\end{table}

\begin{figure*}[t!]
    \centering
    \includegraphics[width=\linewidth]{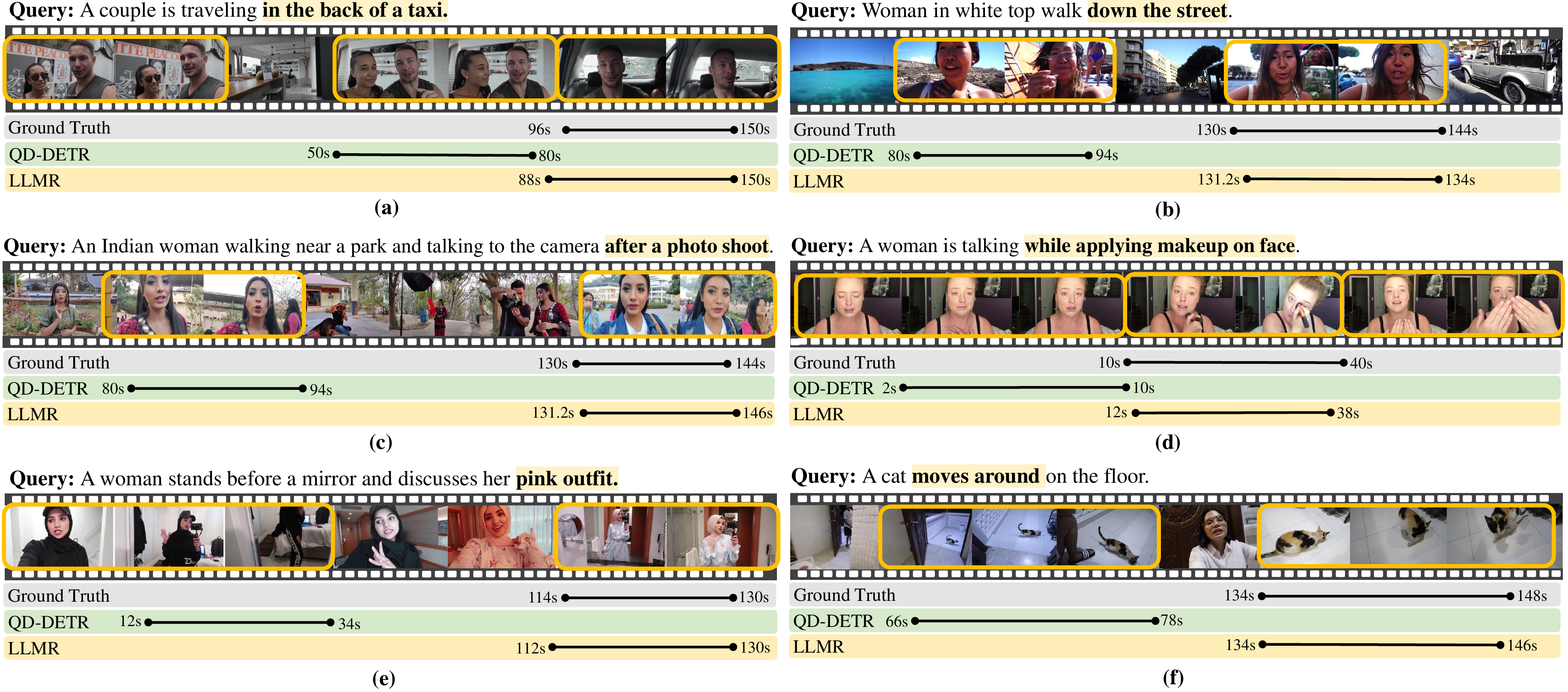}
    \caption{Qualitative results.
    (a)(b): queries with specific backgrounds, (c)(d): queries in specific contextual events, (e): queries with particular appearances, (f): queries specifying certain actions.
    The above constraints are highlighted in yellow in the queries. The yellow rectangular boxes represent moments semantically aligning with the query without yellow-highlighted scene constraints. We observed that QD-DETR~\cite{moon2023query} tends to localize the moments missing the background and action details.
    }
    \label{visualization}
\end{figure*}

\begin{figure}
    \centering
    \includegraphics[width=\linewidth]{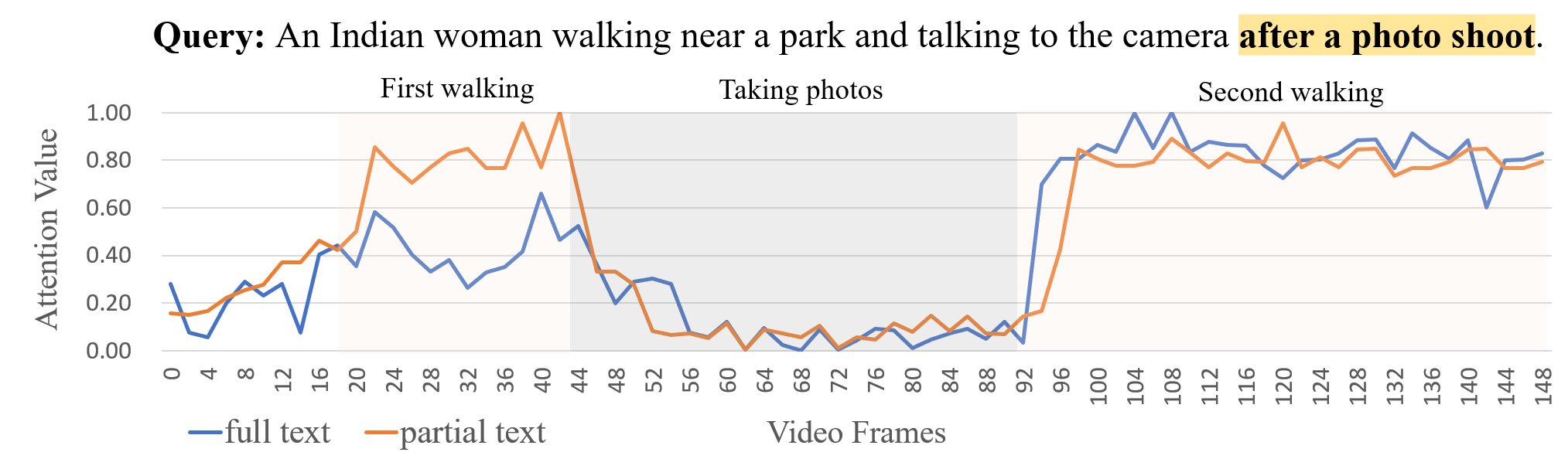}
    \caption{Comparison of attention values with/without specific information.}
    \label{fig}
    \vspace{-11.5pt}
\end{figure}

\vspace{-5pt}
\subsection{Qualitative Analysis}
In \cref{visualization}, we present a qualitative comparison with QD-DETR~\cite{moon2023query}. Without the constraints highlighted in yellow in the queries, each example may encompass multiple matched moments, such as ``a couple is traveling'' (a), ``a woman standing before a mirror'' (e), and ``a cat on the floor''(f). However, when specific constraints are applied, such as backgrounds, contextual events, appearances, and actions, the existing method struggles to distinguish the correct moment from similar-looking and similar-action moments. In contrast, our proposed LMR successfully localizes the target moment. This is attributed to the extensive scene information obtained from Large Language Models (LLMs) for each video clip. When modulated by the query, this information enables the model to dynamically perceive query-relevant detailed changes in scene semantics, such as the color change of character's outfit in (e), the background alteration in (a) and (b), and the subtle movement changes in (d). Consequently, this improves the alignment between complex queries and videos. Additional qualitative examples are provided in the supplementary material.

The \cref{fig} compares the attention values of full-text query and partial-text query by removing ``taking a photo shoot" for example (c) in \cref{visualization}. They are calculated between video context features and text queries.
The lower attention value in ``first walking'' segment in the full-text branch shows that the model truly focuses on the key information in the query text.

\section{Conclusion}
We proposed a Large Language Model-based Moment Retrieval (LMR) approach to tackle the localization challenges posed by complex queries with spatiotemporal context constraints.
We presented a context enhancement technique with the inherent knowledge of LLMs and video-query fusion to generate crucial target-related video context.
We also developed a language-conditioned transformer to directly decode free-form language queries for moment retrieval.
Finally, we evaluated our models on popular VMR benchmarks and our constructed C-QVal, and demonstrated that our approach achieves top-ranked performance.



%
%
\bibliographystyle{IEEEtran}
\bibliography{main}


\newpage

\vspace{11pt}




\vfill

\end{document}